\documentclass[sigconf]{acmart}

\AtBeginDocument{%
  }

\copyrightyear{2024}
\acmYear{2024}
\setcopyright{rightsretained}
\acmConference[KDD '24]{Proceedings of the 30th ACM SIGKDD Conference on Knowledge Discovery and Data Mining}{August 25--29, 2024}{Barcelona, Spain}
\acmBooktitle{Proceedings of the 30th ACM SIGKDD Conference on Knowledge Discovery and Data Mining (KDD '24), August 25--29, 2024, Barcelona, Spain}\acmDOI{10.1145/3637528.3671606}
\acmISBN{979-8-4007-0490-1/24/08}

\usepackage{setspace}
\usepackage{balance} 
\usepackage{multirow}
\usepackage{graphicx}
\usepackage{subcaption}
\usepackage{cleveref}

\usepackage{algorithm}               
\usepackage{algpseudocode}
\usepackage{enumitem}
\usepackage{color, colortbl}

\begin{document}

\title{An Offline Meta Black-box Optimization Framework for Adaptive Design of Urban Traffic Light Management Systems}

\author{Taeyoung Yun}\authornotemark[1]
\affiliation{
  \institution{KAIST}
  \city{Daejeon}
  \country{Republic of Korea}
  \authornote{Both authors contributed equally to this research.}
  }
\email{99yty@kaist.ac.kr}

\author{Kanghoon Lee}\authornotemark[1]
\affiliation{
  \institution{KAIST}
  \city{Daejeon}
  \country{Republic of Korea}
  }
\email{leehoon@kaist.ac.kr}

\author{Sujin Yun}
\affiliation{
  \institution{KAIST}
  \city{Daejeon}
  \country{Republic of Korea}
  }
\email{yunsj0625@kaist.ac.kr}

\author{Ilmyung Kim}
\affiliation{
  \institution{Korea Telecom}
  \city{Seoul}
  \country{Republic of Korea}
}
\email{kim.ilmyung@kt.com}

\author{Won-Woo Jung}
\affiliation{
  \institution{Korea Telecom}
  \city{Seoul}
  \country{Republic of Korea}
}
\email{jung.wonwoo@kt.com}

\author{Min-Cheol Kwon}
\affiliation{
  \institution{Korea Telecom}
  \city{Seoul}
  \country{Republic of Korea}
}
\email{tony.kwon@kt.com}

\author{Kyujin Choi}
\affiliation{
  \institution{Korea Telecom}
  \city{Seoul}
  \country{Republic of Korea}
}
\email{kyujin.choi@kt.com}

\author{Yoohyeon Lee}
\affiliation{%
  \institution{Korea Telecom}
  \city{Seoul}
  \country{Republic of Korea}
}
\email{yoohyeon.lee@kt.com}

\author{Jinkyoo Park}\authornotemark[2]
\affiliation{
  \institution{KAIST}
  \city{Daejeon}
  \country{Republic of Korea}
  \authornote{Corresponding author.}}
\email{jinkyoo.park@kaist.ac.kr}

\renewcommand{\shortauthors}{Taeyoung Yun et al.}

\begin{abstract}
Complex urban road networks with high vehicle occupancy frequently face severe traffic congestion. Designing an effective strategy for managing multiple traffic lights plays a crucial role in managing congestion. However, most current traffic light management systems rely on human-crafted decisions, which may not adapt well to diverse traffic patterns. In this paper, we delve into two pivotal design components of the traffic light management system that can be dynamically adjusted to various traffic conditions: phase combination and phase time allocation. While numerous studies have sought an efficient strategy for managing traffic lights, most of these approaches consider a fixed traffic pattern and are limited to relatively small road networks. To overcome these limitations, we introduce a novel and practical framework to formulate the optimization of such design components using an offline meta black-box optimization. We then present a simple yet effective method to efficiently find a solution for the aforementioned problem. In our framework, we first collect an offline meta dataset consisting of pairs of design choices and corresponding congestion measures from various traffic patterns. After collecting the dataset, we employ the Attentive Neural Process (ANP) to predict the impact of the proposed design on congestion across various traffic patterns with well-calibrated uncertainty. Finally, Bayesian optimization, with ANP as a surrogate model, is utilized to find an optimal design for unseen traffic patterns through limited online simulations. Our experiment results show that our method outperforms state-of-the-art baselines on complex road networks in terms of the number of waiting vehicles. Surprisingly, the deployment of our method into a real-world traffic system was able to improve traffic throughput by 4.80\% compared to the original strategy.

\end{abstract}



\maketitle

\section{Introduction}

Traffic congestion is a growing problem caused by the rise in the number of vehicles and the complexity of urban road networks, leading to wasted time and harmful emissions \cite{bharadwaj2017impact}. In highly developed cities with intricate road networks, solving congestion by modifying or extending the existing traffic infrastructure is expensive and time-consuming. In this context, an intelligent traffic light management system emerges as a promising method to alleviate congestion in urban areas \citep{jahnavi2021intelligent}.

\begin{figure}[t]
    \centering    
    \includegraphics[width=0.95\linewidth]{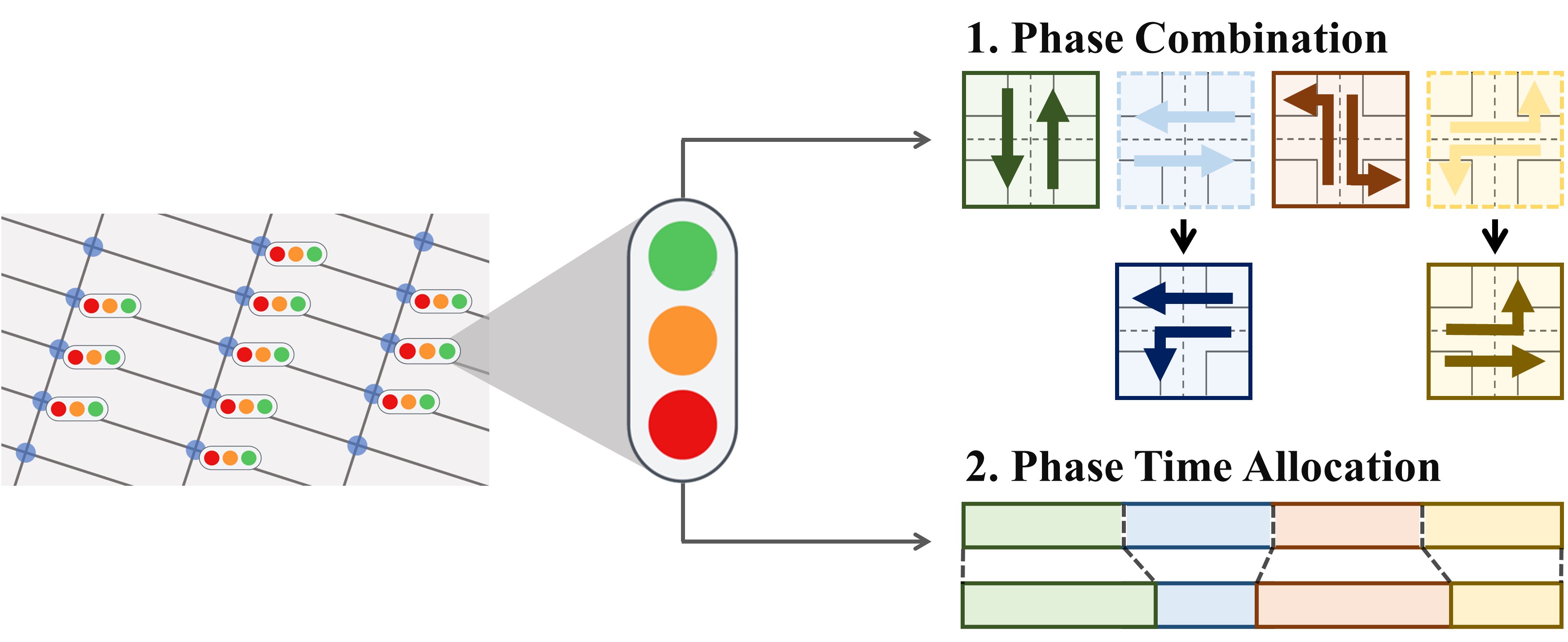}
    \vspace{-7pt}
    \caption{Design components of traffic light scheme.}
    \label{fig:design_illustration}
    \vspace{-15pt}
\end{figure}

Several components of the traffic light management system can be adjusted to significantly mitigate overall road network congestion. 
\Cref{fig:design_illustration} illustrates the two primary components of a traffic light management system that can be optimally designed considering diverse traffic patterns. First, we can determine which combination of phases should be assigned to each traffic light. After selecting the combination, we decide the proportion allocated to each phase (also known as the green split). Both components are crucial for coordinating multiple traffic lights to reduce congestion.

Many traffic engineering methods have been proposed to derive an efficient strategy for managing traffic lights on urban road networks. Classical methods attempt to formulate the green split optimization as mathematical programming, like mixed integer programming. They often rely on strong assumptions such as uniform vehicle arrival rates and fixed right-turn ratios \citep{webster1958traffic, little1981maxband, roess2004traffic, varaiya2013max}. While these methods have shown promising results \citep{lowrie1990scats}, they focus only on single or axis-aligned intersections, and their assumptions may not match well with complex traffic networks. Recent approaches formulate the problem as a black-box optimization and apply various algorithms, including genetic algorithms \citep{zhang2010traffic, li2018signal}, particle swarm optimization \citep{dong2010urban, dabiri2016arterial}, and Bayesian optimization \citep{ito2019coordinated, tay2022bayesian}. 
However, these methods typically solve the problem under a fixed traffic pattern, necessitating an optimization from scratch when a new traffic pattern emerges.
To overcome the aforementioned limitations, we present a novel and practical framework that can adaptively find an optimal design of phase combination and time allocation for diverse traffic patterns. We treat the optimization of such design components as an offline meta black-box optimization. Within this framework, our aim is to find an optimal design for an unseen traffic pattern with a limited number of online simulations while leveraging prior knowledge from a previously collected offline meta dataset.

We propose a simple yet effective method to solve the formulated problem. Our key idea is a novel combination of Attentive Neural Process (ANP) and Bayesian optimization (BO). Initially, we gather an offline meta dataset comprising pairs of design choices and their corresponding traffic measures from various traffic patterns. After collecting the dataset, we train ANP to predict the performance of the proposed design on different traffic patterns with well-calibrated uncertainty, which is crucial in the offline meta-training process.
Subsequently, we integrate the trained ANP model with BO as a means of surrogate model. This integration allows us to effectively utilize the capability of performance prediction on various traffic patterns with well-calibrated uncertainty of the ANP model, enabling us to find an optimal design for unseen traffic patterns with limited online evaluations. 

Our method exhibits superior performance across diverse road networks with realistic traffic patterns.
Moreover, we also deploy our method into real-world traffic light management systems on urban road networks and demonstrate that our method improves traffic flow compared to the original strategy. 
\section{Preliminaries}

\subsection{Traffic Terminology}
In this section, we define some terminologies regarding traffic lights and road network systems.

\textit{Definition 2.1.1 (Traffic Network and Traffic Lights)}. Traffic network can be defined as a graph $\mathcal{G}=(\mathcal{V}, \mathcal{E})$, where $\mathcal{V}$ is a set of intersections and $\mathcal{E}$ is a set of lanes that connect adjacent intersections. Each intersection is equipped with a traffic light device that determines the order and direction of vehicle movements in all lanes at the intersection.

\textit{Definition 2.1.2 (Traffic Pattern)}: Traffic pattern $p$ can be defined as a set of vehicle movements, $\{(o_1, d_1, t_1), (o_2, d_2, t_2), \cdots, (o_V, d_V, t_V)\}$, where $(o_v, d_v)$ represents an origin-destination (OD) pair and $t_v$ denotes the departure time of the $v$th vehicle.

\textit{Definition 2.1.3 (Traffic Light Phase)}. Traffic light phase refers to a specific set of traffic movements that are allowed. During a specific phase, only certain vehicular or pedestrian actions are permitted. Common movements include going straight, turning left, and valid combinations of various movements.

\textit{Definition 2.1.4 (Traffic Light Phase Combination)}. Traffic signal phase combination comprises a set of traffic signal phases. Most real-world traffic lights determine phase combination as 4$\sim$5 number of signal phases and execute them in an iterative manner. 

\textit{Definition 2.1.5 (Traffic Light Phase Time Allocation)}. Once the traffic light determines the phase combination to circle, we can decide the proportion of time allocated to each phase given a fixed cycle time. Phase time allocation refers to such ratio, and it is also often called a green split in transportation literature \citep{ferreira2011impact}.

\subsection{Black-box Optimization}
Black-box optimization (BBO) has become a de-facto framework to formulate the optimization problem where the oracle function is non-convex and non-differentiable. Given a black-box function $f:\mathbb{R}^{d}\rightarrow\mathbb{R}$, we can formulate the problem as follows:
\begin{align}
    \text{find }\mathbf{x}^{*}=\arg\max_{\mathbf{x}\in\mathbb{R}^{d}}f(\mathbf{x})
\end{align}
where $\mathbf{x}$ is our $d$-dimensional decision variable. Numerous methods have been discussed to solve the black-box optimization problem, such as evolutionary algorithms \citep{hansen2006cma}, genetic algorithms \citep{such2017deep}, and Bayesian optimization \citep{frazier2018tutorial}.

In many real-world applications, we constantly encounter new optimization problems that are different but similar to previously seen problems. Unfortunately, naive BBO algorithms start the optimization process from scratch when encountering new problems. To mitigate this issue, researchers turn their eyes to leveraging prior knowledge to accelerate the optimization process for the new black-box function \citep{feurer2018scalable, perrone2018scalable, wistuba2021few}. This approach is termed a meta black-box optimization (Meta BBO), formulated as follows:
\begin{align}
     \text{find }\mathbf{x}^{*}=\arg\max_{\mathbf{x}\in\mathbb{R}^{d}}f(\mathbf{x}),\quad\quad
     \text{Given }f^{1},f^{2}.\cdots,f^{N}
\end{align}
where $f^{1}, f^{2},\cdots,f^{N}$ and $f$ are i.i.d samples from the distribution over functions $\rho(\cdot)$.

In this paper, we present a more practical framework, an offline meta black-box optimization, where we have access only to an offline dataset collected from different black-box functions and online interactions with functions unavailable. In the next section, we will explore our formulation in greater detail.

\subsection{Neural Processes}
Neural processes (NP, \citep{garnelo2018neural}) are the family of stochastic processes, which learn a distribution over functions and have a broad range of applications, including regression \citep{kim2019attentive}, classification \citep{gordon2019convolutional}, and generation tasks \citep{garcia2022conditional}. Unlike Gaussian Processes (GP, \citep{rasmussen2006gaussian}), which have a pre-defined prior for modeling a target function, NP can learn a data-driven prior, enabling them to represent a wide range of complex distributions of functions. 

Formally, NP introduces a latent variable $\mathbf{z}$ and uses neural networks to model encoder and decoder networks. Let $\phi$ and $\theta$ be parameters of the encoder and decoder of NP, respectively. Given context set $\mathcal{D}_{\text{ctx}}=\{(\mathbf{x}_{c}, y_{c})\}_{c=1}^{|C|}$ and target set $\mathcal{D}_{\text{tar}}=\{(\mathbf{x}_{t}, y_{t})\}_{t=|C|+1}^{|C|+|T|}$, the NP encoder first learns a representation $\mathbf{r}_{c}=E_{\phi}(\mathbf{x}_{c}, y_{c})$ for each context. These representations are aggregated to obtain a global representation $\mathbf{r}_{\text{ctx}}$, which is used for parameterizing the distribution of a latent variable $\mathbf{z}\sim q_{\phi}(\mathbf{z}\vert \mathbf{r}_{\text{ctx}})$. Finally, given the sampled latent variable $\mathbf{z}$, the NP decoder returns the predictive distribution of the target set as follows: 
\begin{align}\label{eq:log_likelihood}
    \prod_{t=|C|+1}^{|C|+|T|}p_{\theta}(y_{t}\vert\mathbf{x}_{t}, \mathbf{z})=\prod_{t=|C|+1}^{|C|+|T|}\mathcal{N}(y_{t}\vert\mu_{\theta}(\mathbf{x}_{t}, \mathbf{z}), \sigma^{2}_{\theta}(\mathbf{x}_{t}, \mathbf{z}))
\end{align}

We train the parameters of NP to maximize the target log-likelihood. As it is intractable, we try to maximize the ELBO term below:


\begin{align}
&\log \prod_{t=|C|+1}^{|C|+|T|} p(y_{t}\vert\mathbf{x}_{t},\mathcal{D}_{\text{ctx}})=\log\int \prod_{t=|C|+1}^{|C|+|T|} p(y_{t}\vert\mathbf{x}_{t}, \mathbf{z})p(\mathbf{z}\vert\mathcal{D}_{\text{ctx}})d\mathbf{z}\\
& \geq\mathbb{E}_{q_{\phi}(\mathbf{z}|\mathcal{D}_{\text{tar}})}\left[\sum_{t=|C|+1}^{|C|+|T|}\log p_{\theta}(y_t\vert\mathbf{x}_{t}, \mathbf{z})\right]- D_{KL}\left(q_{\phi}(\mathbf{z}|\mathcal{D}_{\text{tar}})||q_{\phi}(\mathbf{z}|\mathcal{D}_{\text{ctx}})\right)\label{eq:elbo}
\end{align}

A common extension of NP is an Attentive Neural Process (ANP, \cite{kim2019attentive}), which enhances the expressive power of NP. ANP introduces a self-attention module to model higher-order interactions between context pairs while preserving permutation invariance. To predict the target output, ANP employs a cross-attention module instead of mean aggregation so that the target input attends more closely to context representations relevant to the prediction.

\subsection{Bayesian Optimization}
Bayesian optimization (BO) is one of the popular techniques to solve black-box optimization. BO is mainly composed of two parts: surrogate model and acquisition function. The surrogate model approximates the objective function, providing a predictive output value distribution. GP is the most commonly employed surrogate model. 
Then, the acquisition function is utilized to choose the next candidate to evaluate by balancing the predicted mean and variance from the surrogate model. Several acquisition functions have been proposed, including upper confidence bound (UCB, \citep{srinivas2009gaussian}), expected improvement (EI, \citep{movckus1975bayesian}), and probability of improvement (PI, \citep{kushner1964new}). After evaluating the candidate, BO utilizes this new evaluation to update the surrogate function and repeat the process until an optimum is achieved.

\section{Proposed Framework}

This section introduces a novel framework to address the adaptive design optimization problem for deriving an efficient urban traffic light management system. While we focus on the phase time allocation optimization problem to explain our framework, it can be easily extended to the phase combination optimization problem. Please refer to \Cref{app:task_details} for a more detailed task description.

\subsection{Notation}
We begin by defining the decision variable and objective function utilized in our framework.

\textit{Definition 3.1.1} (Decision Variable) Given a traffic network $\mathcal{G}$ with $I$ intersections, we can define our decision variable $\mathbf{x}=[x_{1}, x_{2},\cdots,x_{I}]$ representing our phase time allocation design, where $x_{i}$ denotes phase time allocation design for the $i$th intersection. Specifically, $x_i$ can be defined as a vector, $[x_{i1}, \cdots, x_{iJ}]$, where $J$ is the number of phases and $x_{ij}\in[0, 1]$ is the unnormalized ratio assigned to $j$th phase of the $i$th intersection.

\textit{Definition 3.1.2} (Objective Function). Given a traffic network $\mathcal{G}$ with $I$ intersections and a proposed phase time allocation design $\mathbf{x}$, our objective function $f$ is a mapping from the chosen design $\mathbf{x}$ to the congestion measure. This measure is specified by deploying such design into the traffic system under traffic pattern $p$. 

Congestion measure could be a negative of the average number of waiting vehicles or traffic throughput. As such metrics have no analytically closed form, we can only observe the output $y=f(\mathbf{x}) + \epsilon$ corresponding to the given input, with other intrinsic information remaining hidden. It is a common black-box optimization setting where we should iteratively explore the domain space through trial and error until achieving a satisfactory solution.

\begin{figure*}[t]
    \centering
    \includegraphics[width=0.8\textwidth]{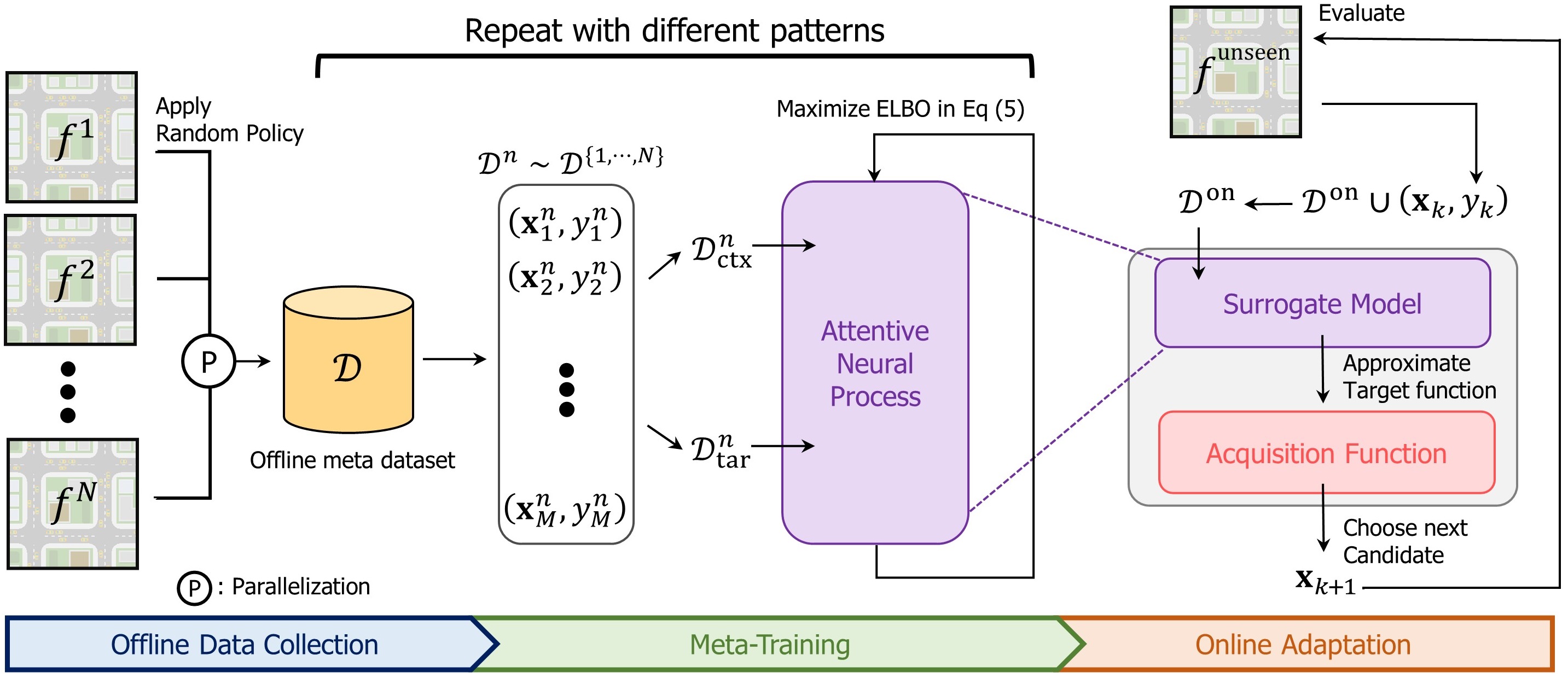}
    \vspace{-7pt}
    \caption{Overview of our proposed method.}
    \vspace{-5pt}
    \label{fig:overview}
\end{figure*}

\subsection{Offline Meta Dataset}
Now we define offline meta dataset $\mathcal{D}$, which consists of input-output pairs from various objective functions. We can readily encounter situations where offline meta dataset exists in various industrial systems. In the context of traffic light systems, we constantly encounter different traffic patterns every day, and we can collect log data of vehicle movements corresponding to various designs using loop detectors or surveillance cameras.

\textit{Definition 3.2.1} (Offline Meta Dataset). Given a traffic network $\mathcal{G}$ with $I$ intersections and set of $N$ objective functions $\{f^{1}, f^{2}, \cdots, f^{N}\}$ specified by different traffic patterns $\{p^{1},p^{2},\cdots,p^{N}\}$, we can collect an offline meta dataset $\mathcal{D}$ consisting of $M$ pairs of designs and the congestion measures for $N$ different functions.
\begin{align}\label{eq:off_dataset}
    \mathcal{D}&=\{\mathcal{D}^{1}, \mathcal{D}^{2}, \cdots, \mathcal{D}^{N}\}\\
    \mathcal{D}^{n}&=(\mathbf{X}^{n}, Y^{n})=\{(\mathbf{x}_{m}^{n}, y_{m}^{n})\}_{m=1}^{M}\quad\forall{n}=1,2,\cdots,N
\end{align}
where $\mathbf{x}_{m}^{n}$ is the $m$th proposed design evaluated on $f^{n}$ and $y_{m}^{n}$ is a corresponding congestion measure, $y_{m}^{n}=f^{n}(\mathbf{x}_{m}^{n})+\epsilon$.

It is natural to assume that underlying similarities exist between different traffic patterns. We believe that exploiting the knowledge from the existing offline meta dataset can help to optimize high-dimensional design problems associated with devising efficient traffic light schemes for large traffic networks.

\subsection{Offline Meta Black-box Optimization}
Finally, we formulate the adaptive design of phase time allocation for an unseen traffic pattern as an offline meta black-box optimization problem.

\textit{Definition 3.3.1} (Offline Meta Black-box Optimization for Adaptive Design of Phase Time Allocation for Unseen Traffic Patterns). Given a traffic network $\mathcal{G}$ with $I$ intersections and offline meta dataset $\mathcal{D}$ collected from $\mathcal{G}$ under $N$ different traffic patterns, our problem can be defined as follows:
\begin{align}
    \mathbf{x}^{*}=\arg\max_{\mathbf{x}}f^{\text{unseen}}(\mathbf{x}),\quad\quad\text{Given }\mathcal{D}\text{ from \Cref{eq:off_dataset}}.
\end{align}
where $f^{\text{unseen}}$ is the objective function specified by unseen traffic pattern, which is not included in the offline meta dataset.

\subsection{Differences with RL-based Approaches}
We emphasize that our method focuses on solving design optimization problems for discovering efficient traffic light management systems instead of deriving a real-time feedback controller. While reinforcement learning (RL) has been considered a promising framework to control traffic lights in real-time \cite{genders2016using, wei2018intellilight, mousavi2017traffic, wei2019presslight, wei2019colight, zhang2020generalight, chen2020toward}, to best of our knowledge, those methods have not been deployed for real traffic network systems due to several challenges \citep{qadri2020state, chen2022real}. Firstly, most RL-based methods define an action as selecting the next phase from a pre-defined set of phases with short decision intervals. This approach can lead to frequent and inconsistent phase transitions, potentially confusing drivers \citep{du2023safelight, muller2022safe}. Moreover, they require the frequent acquisition of accurate traffic flow data in real-time, necessitating sophisticated and expensive sensing and computing technologies \citep{garg2022fully}. Our method distinguishes itself from those methods by framing the problem as a static optimization.
\section{Proposed Method}

In this section, we propose a simple yet effective method to solve the problem formulated in the previous section. Our method encompasses three primary stages: Offline Data Collection, Meta Training, and Online Adaptation. We innovate each process to enhance our method. First, we accelerate the data collection process via parallelization with random sampling. Subsequently, we meta-train ANP using the collected offline meta dataset. Lastly, we employ Bayesian optimization for few-shot online adaptation with the trained ANP as a surrogate model. \Cref{fig:overview} provides an overview of our method.

\subsection{Offline Data Collection}
We construct the offline meta dataset by evaluating various designs on different traffic functions. For sampling design $\mathbf{x}$, we use random sampling strategy. While it may be far from the optimal strategy, it is efficient since we can parallelize the data collection process. Conversely, collecting samples via optimization strategies must be done sequentially as they make decisions based on previous evaluations, which is computationally expensive.

\subsection{Meta-Training Neural Process with Offline Data}
We employ ANP to predict the distribution of the objective function $f$ given a context set $\mathcal{D}_{\text{ctx}}=\{(\mathbf{x}_{c}, y_{c})\}_{c=1}^{|C|}$. As a type of meta-learning, ANP can effectively infer the objective function with few evaluations. In addition, its uncertainty quantification helps overcome the distributional shift issue when using the offline dataset.

\Cref{fig:overview} depicts the process of how we train ANP with the offline meta dataset. For each training step, we arbitrarily choose a dataset $\mathcal{D}^{n}$ from the offline meta dataset $\mathcal{D}$. Then, we draw samples from the $\mathcal{D}^{n}$ and split them into context and target sets. 
\begin{align}
    &\mathcal{D}^{n}_{\text{ctx}}=\{(\mathbf{x}^{n}_{c}, y^{n}_{c})\}_{c=1}^{\vert C\vert},\quad\mathcal{D}^{n}_{\text{tar}}=\{(\mathbf{x}^{n}_{t}, y^{n}_{t})\}_{t=\vert C\vert+1}^{\vert C\vert+\vert T\vert}\\
    &\mathcal{D}^{n}_{\text{ctx}},\;\mathcal{D}^{n}_{\text{tar}}\subset\mathcal{D}^{n}
\end{align}
where $\vert C\vert, \vert T\vert$ represent the number of samples in the context and target sets, respectively. We train ANP to maximize the ELBO term defined in \Cref{eq:elbo} for sampled context set $\mathcal{D}^{n}_{\text{ctx}}$ and target set $\mathcal{D}^{n}_{\text{tar}}$.
Training ANP with samples from different traffic patterns enables learning of a data-driven prior, which can be used to quickly infer unseen traffic pattern with a small number of samples acquired from that pattern.


\subsection{Online Adaptation with Bayesian Optimization}
After meta-training ANP, we employ BO to find an optimum design for unseen traffic patterns as shown in \Cref{fig:overview}. Firstly, we set $\mathcal{D}^{\text{on}}$ as an empty set and initialize a surrogate model with a meta-trained ANP. For each evaluation step $k$, we approximate the target function $f^{\text{unseen}}$ with the surrogate model. In other words, our model infers a posterior distribution of the target function conditioned on $\mathcal{D}^{\text{on}}$. ANP projects $\mathcal{D}^{\text{on}}$ into $\mathbf{z}$ and implicitly captures the similarity between the target traffic pattern and patterns that are used for collecting offline meta dataset in the latent space. 

Given posterior distribution, the acquisition function identifies an optimal-looking candidate by balancing the predicted mean and variance. Among various acquisition functions, we use upper confidence bound (UCB) as a default setting. After selecting $\mathbf{x}_{t+1}$, we evaluate the proposed design and get results, $y_{k+1}=f^{\text{unseen}}(\mathbf{x}_{k+1})+\epsilon$. Then we update the online dataset $\mathcal{D}^{on}\gets\mathcal{D}^{on}\cup\{(\mathbf{x}_{k+1}, y_{k+1})\}$ and repeat the process until the end of the budget $K$. 

\section{Experiments}
This section presents our experiment results on adaptive design optimization tasks, phase combination, and phase time allocation. We validate our method on various sizes of traffic networks.

\subsection{Experiment Setting}
We conduct our experiments on SUMO \citep{behrisch2011sumo}, an open-source traffic simulator supporting a large-scale traffic lights management system. For all experiments, we evaluate the proposed design by aggregating the average number of waiting vehicles from all intersections.

\begin{figure}[t]
    \centering    
    \includegraphics[width=0.9\linewidth]{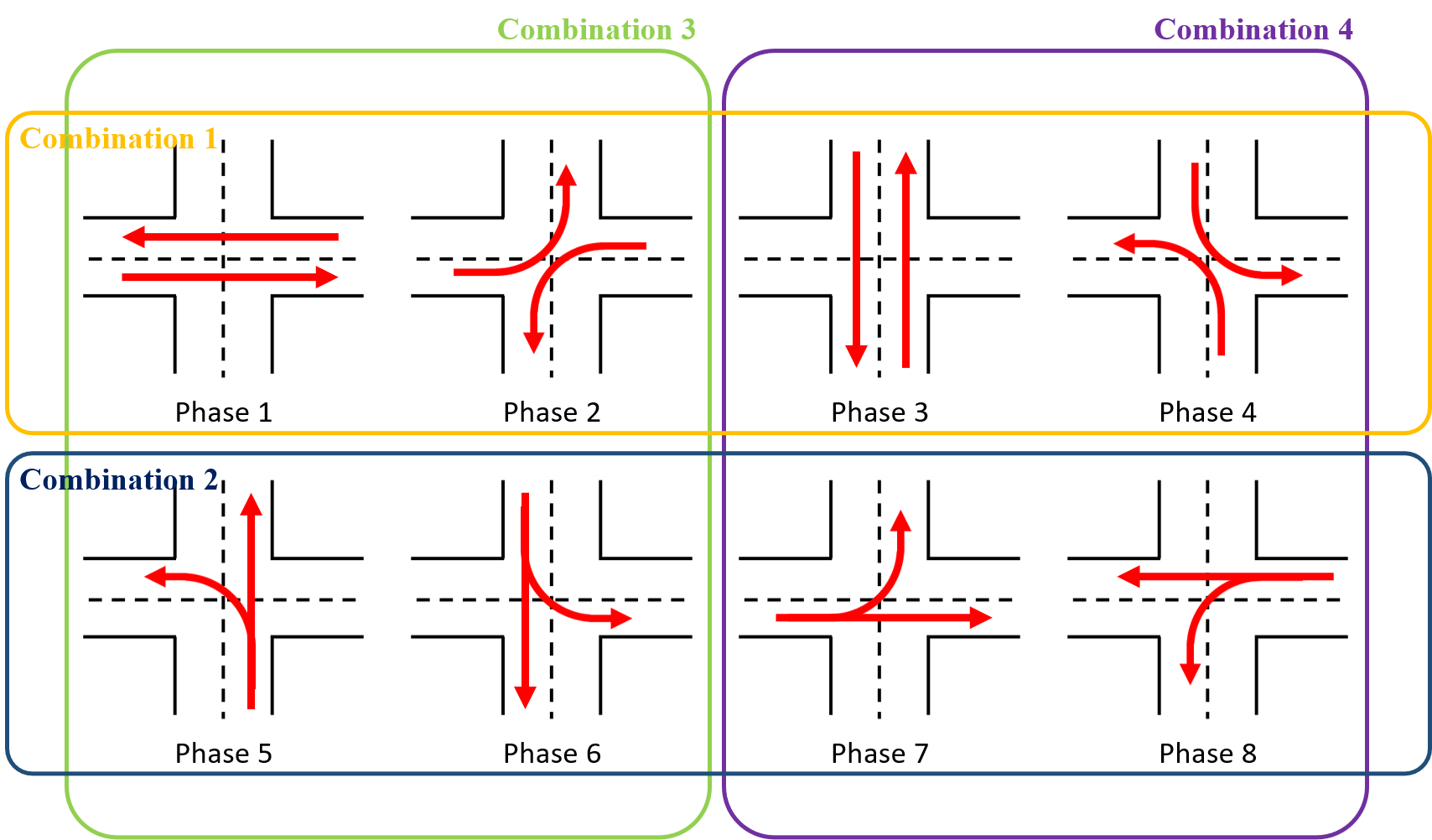}
    \vspace{-7pt}
    \caption{Illustration of possible phases and combinations.}
    \vspace{-10pt}
    \label{fig:phase_comb}
\end{figure}

\subsubsection{Traffic Environments}
\begin{figure}[t]
    \centering    
    \includegraphics[width=0.9\linewidth]{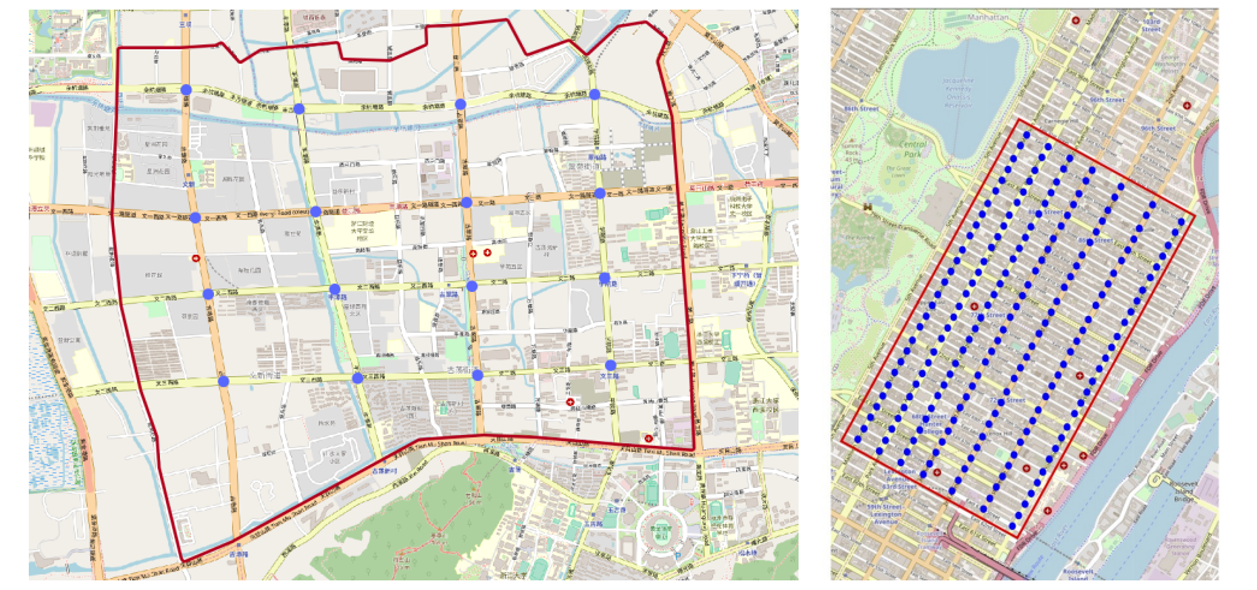}
    \vspace{-7pt}
    \caption{Illustration of road networks used for our experiments: (Left) Hangzhou\_4$\times$4, (Right) Manhattan\_28$\times$7. Figures are taken from \citep{wei2019colight}.}
    \vspace{-10pt}
    \label{fig:network}
\end{figure}

We evaluate our method on seven different traffic road networks, including four synthetic road networks and three real-world road networks, Hangzhou\_4$\times$4, Manhattan\_16$\times$3, and Manhattan\_28$\times$7. \Cref{fig:network} illustrates the real road networks used in our experiments.

\begin{itemize}[leftmargin=1em]
    \item Synthetic road networks: We create four synthetic road networks with different sizes: 2$\times$2, 3$\times$3, 4$\times$4, and 5$\times$5 grids. 
    \item Hangzhou\_4$\times$4: It consists of 16 intersections in Gudang Sub-district with traffic data generated from the surveillance cameras.
    \item Manhattan\_16$\times$3, 28$\times$7: Each network comprises 48 and 196 intersections in the upper east side of Manhattan with traffic data from the open-source taxi trip data. 
\end{itemize}
In case of the synthetic networks, we pre-define a set of origin-destination pairs and sample the interval $\Delta t$ between vehicles for each pair from randomly initialized Poisson distribution to create diverse patterns. For real-world networks, as we already have a predetermined traffic pattern, we randomly reassign departure time $t$ for each vehicle to create diverse traffic patterns. For further details, please refer to \Cref{app:exp_setup}.

\subsubsection{Design Components}
We evaluate our method on two different adaptive design optimization tasks for coordinating multiple traffic lights on urban road networks.
\begin{itemize}[leftmargin=1em]
    \item Phase combination: We choose a phase combination among pre-defined four valid phase combinations for all intersections. \Cref{fig:phase_comb} shows possible combinations of phases. We evenly distribute cycle time into all phases.
    \item Phase time allocation: We choose the ratio to be assigned for each phase at every intersection. We use the Combination 1 in \Cref{fig:phase_comb} as a default setting. We also guarantee at least 30 seconds for all phases. The cycle time is set to 180.
\end{itemize}
Note that we guarantee minimum green time for each phase. While it is a practical assumption since pedestrians require adequate time to pass crosswalks in real-world, it is frequently overlooked in real-time phase controlling strategies.

\subsubsection{Offline Data Collection}
For all experiments, we prepare an offline meta dataset from 120 different traffic patterns and split it into a ratio of 5:1 for train and valid datasets.
We evaluate 200 randomly sampled designs for each traffic pattern in both datasets. In Manhattan\_28$\times$7 network with phase combination design problem, the search space is $4^{28\times7}$, where our offline meta dataset can cover only a tiny fraction of the whole search space.

\begin{table*}[t]
\centering
\caption{Performance of all methods on phase combination task. Mean and one standard deviation over 3 seeds are reported.}
\vspace{-7pt}
\resizebox{0.95\linewidth}{!}{
\begin{tabular}{clcccc|ccc}
\toprule
\multirow{3}{*}{\textbf{Framework}} & \multirow{3}{*}{\textbf{Method}} & \multicolumn{7}{c}{\textbf{Average Number of Waiting Vehicles ($\downarrow$)}}\\ 
\cmidrule{3-9}
& & Grid\_2$\times$2 & Grid\_3$\times$3 & Grid\_4$\times$4 & Grid\_5$\times$5 & Hangzhou\_4$\times$4 & Manhattan\_16$\times$3 & Manhattan\_28$\times$7\\
\midrule
\multirow{5}{*}{BBO} 
& GA & 58.02 ± 0.44 & 147.16 ± 1.99 & 281.00 ± 3.72 & 401.00 ± 16.07 & 294.45 ± 3.25 & 909.25 ± 4.15 & 3667.06 ± 7.57\\
& PSO & 60.65 ± 1.79 & 150.16 ± 0.55 & 281.73 ± 3.72 & 398.50 ± 18.85 & 296.81 ± 1.55 & 909.02 ± 6.95 & 3671.08 ± 0.24\\
& CMA-ES & 58.98 ± 0.82 & 147.56 ± 3.09 & 278.59 ± 8.03 & 376.94 ± 13.15 & 292.85 ± 1.16 & 906.88 ± 0.82 & 3664.38 ± 0.61\\
& GP-UCB & 61.67 ± 1.19 & 160.07 ± 3.35 & 283.74 ± 5.03 & 383.03 ± 7.19 & 295.43 ± 0.00 & 907.53 ± 9.27 & 3675.98 ± 4.53\\
& REINFORCE & 59.69 ± 1.72 & 151.09 ± 11.96 & 289.56 ± 7.37 & 381.84 ± 5.57 & 299.80 ± 5.27 & 916.88 ± 1.15 & 3691.42 ± 1.15\\
\midrule
\multirow{5}{*}{Meta BBO} 
& LGA & 59.08 ± 0.11& 149.47 ± 3.38 & 282.70 ± 2.34 & 386.08 ± 19.35 & 298.60 ± 2.17 & 903.42 ± 4.05 & 3682.88 ± 1.34\\
& LES & 59.97 ± 2.49 & 154.99 ± 1.45 & 282.38 ± 4.53 & 412.61 ± 6.10 & 299.97 ± 3.79 & 918.85 ± 3.82 & 3673.02 ± 2.59\\
& RGPE & 57.53 ± 0.27 & 148.77 ± 1.84 & 281.06 ± 4.41 & 379.39 ± 3.81 & 294.78 ± 2.00 & 910.20 ± 15.87 & 3678.96 ± 6.08\\
& ABLR & 58.52 ± 1.08 & 149.19 ± 1.21 & 282.53 ± 2.64 & 393.76 ± 17.34 & 296.04 ± 1.85 & 907.13 ± 7.60 & 3673.28 ± 0.72\\
& FSBO & 58.50 ± 0.39 & 146.11 ± 1.69 & 283.38 ± 2.24 & 396.28 ± 8.63 & 295.13 ± 1.33 & 900.50 ± 2.60 & 3669.48 ± 7.98\\
\midrule
\multirow{1}{*}{Offline Meta BBO} 
& Ours &\cellcolor{lightgray}\textcolor{black}{\textbf{57.45 ± 0.06}} & \cellcolor{lightgray}\textcolor{black}{\textbf{142.57 ± 3.48}} & \cellcolor{lightgray}\textcolor{black}{\textbf{278.04 ± 1.74}} & \cellcolor{lightgray}\textcolor{black}{\textbf{347.99 ± 4.77}} & \cellcolor{lightgray}\textcolor{black}{\textbf{292.69 ± 1.16}} & \cellcolor{lightgray}\textcolor{black}{\textbf{899.42 ± 9.15}} & \cellcolor{lightgray}\textcolor{black}{\textbf{3657.28 ± 13.72}}\\
\bottomrule
\end{tabular}}
\label{table:exp_comb}
\end{table*}

\begin{table*}[t]
\centering
\caption{Performance of all methods on phase time allocation task. Mean and one standard deviation over 3 seeds are reported.}
\vspace{-7pt}
\resizebox{0.95\linewidth}{!}{
\begin{tabular}{clcccc|ccc}
\toprule
\multirow{3}{*}{\textbf{Framework}} & \multirow{3}{*}{\textbf{Method}} & \multicolumn{7}{c}{\textbf{Average Number of Waiting Vehicles ($\downarrow$)}}\\ 
\cmidrule{3-9}
& & Grid\_2$\times$2 & Grid\_3$\times$3 & Grid\_4$\times$4 & Grid\_5$\times$5 & Hangzhou\_4$\times$4 & Manhattan\_16$\times$3 & Manhattan\_28$\times$7\\
\midrule
\multirow{5}{*}{BBO} 
& GA & 53.39 ± 0.60 & 152.19 ± 0.47 & 261.97 ± 1.56 & 430.13 ± 1.26 & 403.06 ± 1.04 & 904.68 ± 5.70 & 3890.35 ± 0.45\\
& PSO & 54.17 ± 2.60 & 151.99 ± 1.21 & 262.32 ± 1.91 & 428.62 ± 2.09 & 403.20 ± 1.07 & 920.26 ± 8.59 & 3903.77 ± 2.98\\
& CMA-ES & 54.82 ± 0.78 & 151.23 ± 0.46 & 261.61 ± 0.69 & 427.60 ± 1.14 & 401.85 ± 2.23 & 911.06 ± 1.13 & 3886.70 ± 3.61\\
& GP-UCB & 55.14 ± 0.20 & 152.60 ± 0.40 & 260.15 ± 0.73 & 427.71 ± 0.59 & 404.37 ± 0.26 & 916.11 ± 4.00 & 3891.94 ± 3.01\\
& REINFORCE & 50.80 ± 0.98 & 151.86 ± 0.73 & 269.63 ± 0.56 & 433.32 ± 0.09 & 402.53 ± 2.16 & 935.74 ± 1.58 & 3897.91 ± 3.10\\
\midrule
\multirow{5}{*}{Meta BBO} 
& LGA & 51.73 ± 0.67 & 150.58 ± 0.13 & 263.08 ± 4.40 & 426.31 ± 0.47 & 397.04 ± 1.08 & 906.81 ± 7.36 & 3869.06 ± 5.41\\
& LES & 54.67 ± 2.07 & 152.92 ± 2.08 & 263.98 ± 4.20 & 431.86 ± 1.78 & 411.13 ± 5.13 & 930.45 ± 7.11 & 3906.57 ± 12.63\\
& RGPE & 51.40 ± 0.26 & 151.40 ± 0.19 & 261.01 ± 0.91 & 425.93 ± 0.13 & 406.80 ± 1.31 & 918.04 ± 2.04 & 3894.00 ± 2.85\\
& ABLR & 48.58 ± 1.50 & 149.95 ± 0.79 & 259.36 ± 2.36 & 427.64 ± 0.48 & 401.79 ± 3.15 & 916.76 ± 4.07 & 3891.27 ± 3.38\\
& FSBO & 48.40 ± 1.10 & 149.79 ± 0.67 & 259.86 ± 1.79 & 429.71 ± 0.56 & 407.49 ± 0.43 & 913.90 ± 2.25 & 3893.21 ± 1.80\\
\midrule
\multirow{1}{*}{Offline Meta BBO} 
&  Ours & \cellcolor{lightgray}\textcolor{black}{\textbf{46.47 ± 0.55}} & \cellcolor{lightgray}\textcolor{black}{\textbf{146.94 ± 0.97}} & \cellcolor{lightgray}\textcolor{black}{\textbf{247.05 ± 2.48}} & \cellcolor{lightgray}\textcolor{black}{\textbf{421.74 ± 0.51}} & \cellcolor{lightgray}\textcolor{black}{\textbf{393.66 ± 0.42}} & \cellcolor{lightgray}\textcolor{black}{\textbf{850.53 ± 9.63}} & \cellcolor{lightgray}\textcolor{black}{\textbf{3805.36 ± 15.96}}\\
\bottomrule
\end{tabular}}
\label{table:exp_timing}
\end{table*}

\subsection{Baselines}
We consider several state-of-the-art baselines suitable for design optimization problems.
\begin{itemize}[leftmargin=1em]
    \item \textbf{BBO}: Black-box optimization methods search an optimal design by trial and error through iterative online evaluations. Genetic algorithm (GA \citep{such2017deep}), particle swarm optimization (PSO \citep{kennedy1995particle}), GP with UCB as an acquisition function (GP-UCB \citep{srinivas2009gaussian}), evolutionary algorithm (CMA-ES \citep{hansen2006cma}), and reinforcement learning based method (REINFORCE \citep{williams1992simple}) are included.
    \item \textbf{Meta BBO}: We also consider methods that leverage prior knowledge from various tasks to solve the unseen problem more efficiently rather than starting from scratch. We consider LGA \citep{lange2023discovering} and LES \citep{tjarko2022discovering}, which parameterize components in genetic and evolutionary algorithms, respectively. We also include meta Bayesian optimization methods, RGPE \citep{feurer2018scalable}, ABLR \citep{perrone2018scalable}, and FSBO \citep{wistuba2021few} as state-of-the-art baselines.
\end{itemize}

\subsection{Implementation}
For implementing baselines, we try to strictly follow implementations from authors. To implement our method, we use publicly available code\footnote{https://github.com/juho-lee/bnp.git}. We employ multi-layer perceptron (MLP) to parametrize encoder and decoder networks for all experiments. We train the model with 10,000 epochs except for Manhattan scenarios. Since the Manhattan scenarios have very high dimensional input space, we increase the number of epochs to 25,000. For each training step, we utilize a batch size of 16. The size of the context set $|C|$ is sampled from $\text{Unif}(10, 190)$, and the size of the target set $|T|$ is sampled from $\text{Unif}(10, 200-|C|)$. Please refer to \Cref{app:hyperparameters} for more details on implementations.




\subsection{Main Results}
We compare the performance of the proposed method and baselines on different traffic networks. For testing, we sample 10 new traffic patterns, which have never been included in the offline meta dataset. We then perform optimization with $K$ online trials for each traffic pattern and report the average of best performance among trials. For Grid\_2$\times$2, 3$\times$3 and 4$\times$4 networks, we set $K=40, 60$ and $80$ respectively. For other networks, we set $K$ as 100.

\Cref{table:exp_comb} and \Cref{table:exp_timing} summarize the overall experiment results. As indicated in tables, our method consistently outperforms baselines. We find that Meta BBO baselines generally exhibit higher performance compared to other baselines, which have no capability to capture underlying similarities between different traffic patterns. However, those methods do not consistently perform well across different design components and traffic networks, especially when the scale of network grows. The performance gap between GP-UCB and our method also further underscores the superiority of meta-trained ANP as a surrogate model. 

\begin{figure}[t]
    \centering    
    \includegraphics[width=1.0\linewidth]{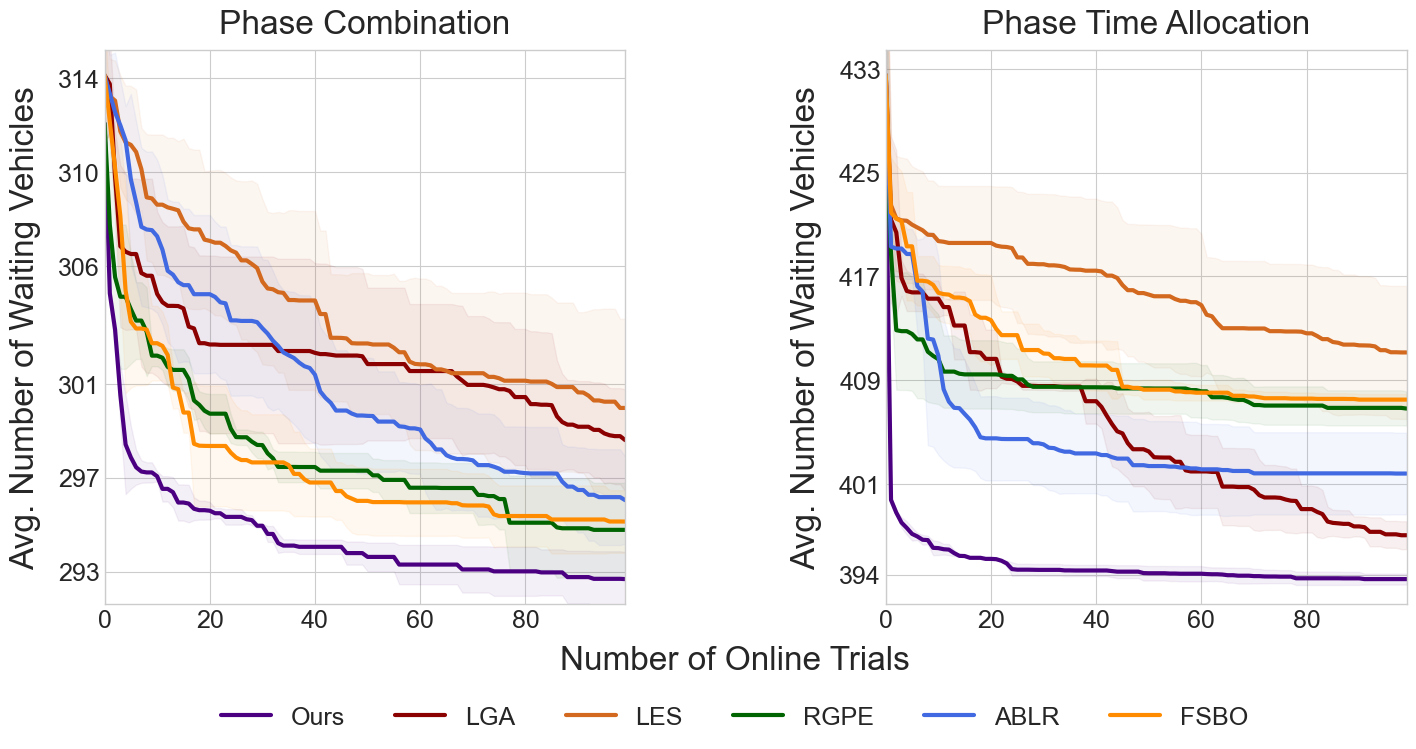}
    \vspace{-18pt}
    \caption{Performance comparison between meta BBO methods and our method on Hangzhou Network across online trials. Mean and one standard deviation across three random seeds are reported.}
    \label{fig:hangzhou_main}
    \vspace{-10pt}
\end{figure}

\Cref{fig:hangzhou_main} (Left) displays the performance of our proposed method and baselines over the course of evaluation for phase combination task on the Hangzhou network. As depicted in the figure, our method not only achieves superior performance but also converges more rapidly to the high-performing solution. These results support our claim that ANP can discern underlying prior distribution over different traffic patterns even when trained on offline meta dataset. Then BO utilizes that information from ANP to search the domain space efficiently, which synergistically leads to superior performance. \Cref{fig:hangzhou_main} (Right) shows the performance of our method and baselines for the phase time allocation task on Hangzhou network. We also observe a similar trend on phase time allocation task, showcasing the efficacy of our method across different design components. 

\begin{figure}[t]
    \centering
    \includegraphics[width=0.8\columnwidth]
    {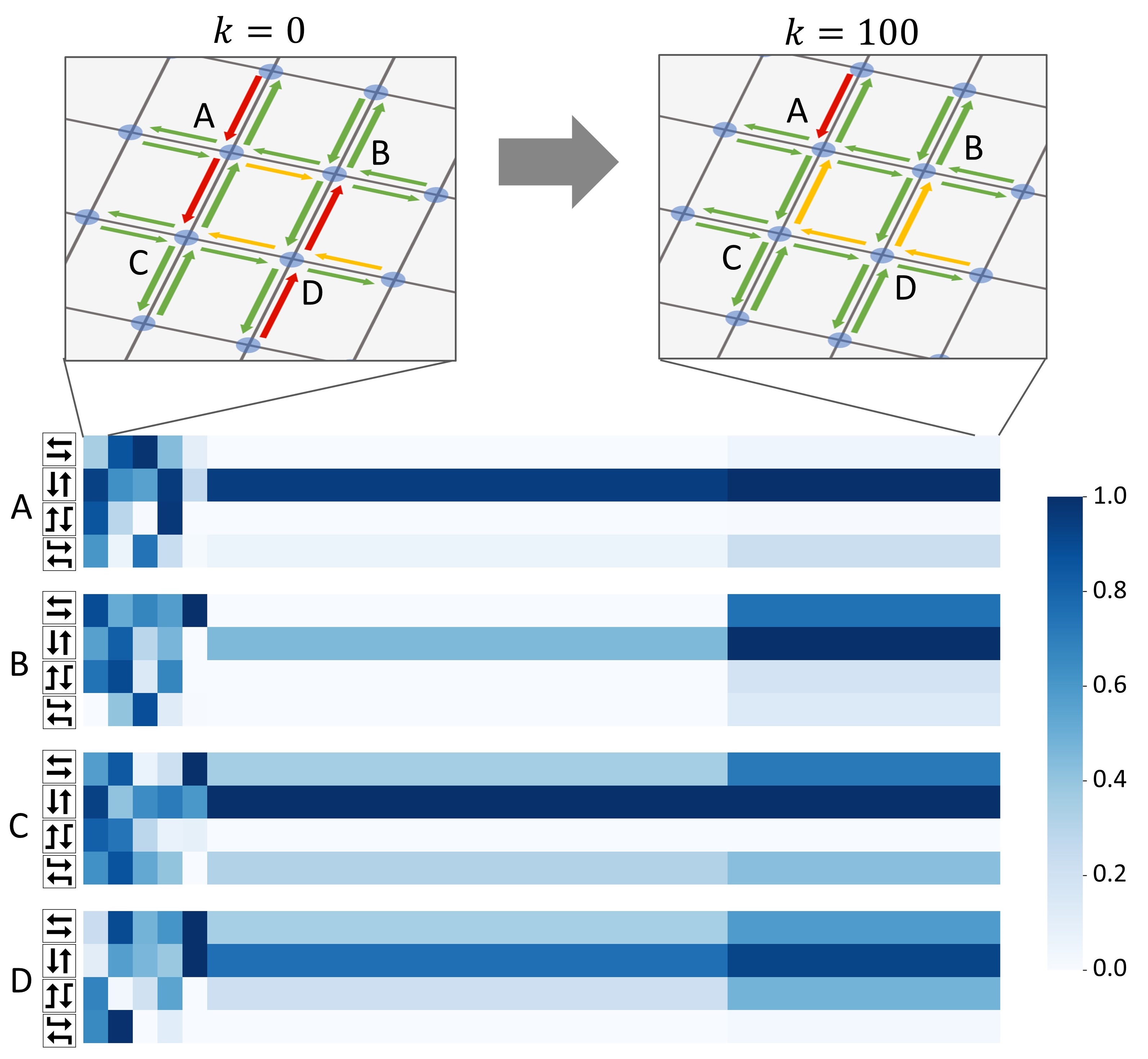}
    \vspace{-7pt}
    \caption{The optimization process of our method for the sub-region of the Manhattan\_16$\times$3. Heatmaps show the best phase time allocation across BO iterations.}
    \label{fig:opt_results}
    \vspace{-8pt}
\end{figure}

\subsection{Qualitative Analysis}
We analyze the efficiency of our proposed method in determining the appropriate design for unseen traffic patterns adaptively by comparing congestion levels in the Manhattan\_16$\times$3 network before and after optimization. \Cref{fig:opt_results} represents the congestion level of each lane in the sub-region. The lines in green, yellow, and red, respectively, denote light, moderate, and heavy traffic conditions. As evident from the figure, our approach significantly reduces congestion. We also illustrate the evolution of the phase time allocation using heatmap, where a deeper block indicates longer time is allocated to the phase. As observed, after a few online adaptation steps, our model recommends an extended green light duration for the North$\leftrightarrow$South direction to all intersections. This strategy aligns well with the given traffic flow pattern, where a major traffic volume concentrates on North$\leftrightarrow$South direction.




\subsection{Comparison with RL-based Methods}
In this section, we discuss the difference between our framework
with RL-based methods. To facilitate this, we compare the performance of RL-based methods on phase combination optimization tasks. For a fair comparison, we formulate the problem as a 1-step MDP and provide only a single global reward at the terminal step. This is a practical setting since on-the-fly traffic information gathering and processing in real-world traffic systems is costly.

We consider several baselines, including Q-Learning based and Actor-Critic methods. We adopt code from LibSignal \citep{mei2022libsignal}, an open library for RL-based traffic signal control. DQN \citep{mnih2015human, van2016coordinated} and PPO \citep{schulman2017proximal} train single central RL policy to govern all intersections, while independent DQN (IDQN, \citep{zheng2019diagnosing}) and independent PPO (IPPO) consider each intersection as an independent RL agent. Finally, we also compare the performance of recent RL-based methods tailored for traffic light control, CoLight \citep{wei2019colight} and MAPG \citep{chu2019multi}. Both methods utilize neighborhood information to promote the coordination of multiple traffic lights.

\Cref{table:rl_comparison} shows the performance of RL-based methods. While recent RL-based methods such as CoLight and MAPG show relatively good performance than other baselines, our method still surpasses those methods. It seems that coordinating multiple traffic lights using a global reward signal is notoriously difficult for RL-based methods, especially under limited online evaluations \citep{wang2019influence}.


\begin{table}[t]
\centering
\caption{Comparison with RL-based methods on phase combination optimization task.}
\vspace{-7pt}
\resizebox{\linewidth}{!}{
\begin{tabular}{ccccc}
\toprule
\multirow{2}{*}{\textbf{Method}} & \multicolumn{4}{c}{\textbf{Average Number of Waiting Vehicles ($\downarrow$)}}\\ 
\cmidrule{2-5}
& Grid\_2$\times$2 & Grid\_3$\times$3 & Grid\_4$\times$4 & Grid\_5$\times$5 \\
\midrule
DQN & 60.17 ± 2.24 & 156.58 ± 0.58 & 295.87 ± 3.87 & 404.38 ± 0.82 \\
PPO & 59.33 ± 0.98 & 153.17 ± 1.20 & 290.20 ± 3.86 & 402.02 ± 0.43 \\
IDQN & 60.51 ± 0.42 & 154.09 ± 4.78 & 293.42 ± 3.44 & 400.99 ± 12.13 \\
IPPO & 58.96 ± 1.41 & 154.84 ± 0.17 & 293.71 ± 1.21 & 397.21 ± 0.79 \\
CoLight & 59.33 ± 0.85 & 153.55 ± 4.16 & 292.69 ± 3.51 & 392.41 ± 14.77 \\
MAPG & 58.36 ± 1.33 & 153.90 ± 1.20 & 290.93 ± 2.52 & 396.63 ± 5.71 \\
\midrule
Ours & \cellcolor{lightgray}\textcolor{black}{\textbf{57.45 ± 0.06}} & \cellcolor{lightgray}\textcolor{black}{\textbf{142.57 ± 3.4}} & \cellcolor{lightgray}\textcolor{black}{\textbf{278.04 ± 1.74}} & \cellcolor{lightgray}\textcolor{black}{\textbf{347.99 ± 4.77}}\\
\bottomrule
\end{tabular}}
\label{table:rl_comparison}
\vspace{-5pt}
\end{table}

\subsection{Additional Experiments}
In this section, we present further experiments to deepen our understanding of the proposed method.


\begin{table}[t]
\centering
\caption{Ablation on the size of offline meta dataset. Experiment is conducted on Grid\_3$\times$3 phase time allocation task.}
\vspace{-2pt}
\resizebox{1.0\linewidth}{!}{
\begin{tabular}{ccc|cc}
\toprule
\multirow{3}{*}{\textbf{Method}} & \multicolumn{4}{c}{\textbf{Average Number of Waiting Vehicles ($\downarrow$)}}\\ 
\cmidrule{2-5}
& \multicolumn{2}{c}{Number of Tasks} & \multicolumn{2}{c}{Number of Samples}\\
\cmidrule{2-5}
& N20-M200 & N50-M200 & N100-M40 & N100-M100\\
\midrule
RGPE & 150.45 ± 0.41 & 150.95 ± 0.12 & 151.86 ± 0.04 & 151.26 ± 0.22 \\
ABLR & 150.68 ± 0.94 & 147.76 ± 0.47 & 150.44 ± 0.32 & 148.07 ± 0.43 \\
FSBO & 150.16 ± 0.19 & 150.83 ± 0.11 & 150.06 ± 0.64 & 150.96 ± 0.12 \\
\midrule 
Ours & \cellcolor{lightgray}\textcolor{black}{\textbf{147.15 ± 0.27}} & \cellcolor{lightgray}\textcolor{black}{\textbf{147.15 ± 0.27}} & \cellcolor{lightgray}\textcolor{black}{\textbf{147.80 ± 0.59}} & \cellcolor{lightgray}\textcolor{black}{\textbf{146.89 ± 1.24}} \\
\bottomrule
\end{tabular}
}
\vspace{-10pt}
\label{table:ablation1}
\end{table}

\subsubsection{\textbf{Size of the Offline Meta Dataset}}\label{app:add_exp1}
It is important to check the robustness of performance in terms of the size of the dataset since we cannot collect infinitely large amounts of data in real-world traffic scenarios. To address this, we conduct additional experiments by varying the size of the offline meta dataset in two ways: (1) changing the number of different traffic patterns, $N$, and (2) changing the number of input-output pairs per pattern, $M$. We perform experiments on Grid 3$\times$3 network with phase time allocation optimization task.

We compare the performance of our method with baselines, RGPE, ABLR, and FSBO, which utilize the offline meta dataset. The results from these experiments are listed in \Cref{table:ablation1}. As the table illustrates, our model outperforms other baselines regardless of the number of traffic patterns and samples collected per pattern. 
It suggests our model effectively identifies hidden similarities between different traffic patterns even with the small size of the dataset.
We also find that the performance of our model constantly improves as the scale of the dataset increases, showcasing its capability.


\begin{table}[t]
\centering
\caption{Experiment Results on Different Evaluation Metrics. Experiment is conducted on phase time allocation task}
\vspace{-2pt}
\resizebox{1.0\linewidth}{!}{
\begin{tabular}{ccccc}
\toprule
\multirow{2}{*}{\textbf{Method}} & \multicolumn{4}{c}{\textbf{Average Traveling Time (sec, $\downarrow$)}}\\ 
\cmidrule{2-5}
& Grid\_2$\times$2 & Grid\_3$\times$3 & Grid\_4$\times$4 & Grid\_5$\times$5 \\
\midrule
RGPE & 202.93 ± 1.37 & 373.86 ± 0.58 & 473.06 ± 1.12 & 634.46 ± 0.17 \\
ABLR & 194.39 ± 0.45 & 370.56 ± 0.75 & 466.40 ± 0.08 & 630.39 ± 0.95 \\
FSBO & 194.72 ± 2.44 & 372.54 ± 0.05 & 473.06 ± 1.12 & 634.71 ± 0.25 \\
\bottomrule
Ours & \cellcolor{lightgray}\textcolor{black}{\textbf{187.36 ± 0.54}} & \cellcolor{lightgray}\textcolor{black}{\textbf{365.97 ± 2.08}} & \cellcolor{lightgray}\textcolor{black}{\textbf{450.94 ± 5.74}} & \cellcolor{lightgray}\textcolor{black}{\textbf{617.00 ± 1.85}} \\
\bottomrule
\end{tabular}
}
\label{table:ablation2_1}
\end{table}
\begin{table}[t]
\centering
\caption{Experiment Results on Different Evaluation Metrics. Experiment is conducted on phase time allocation task}
\vspace{-2pt}
\resizebox{1.0\linewidth}{!}{
\begin{tabular}{ccccc}
\toprule
\multirow{2}{*}{\textbf{Method}} & \multicolumn{4}{c}{\textbf{CO2 Emissions (g/sec, $\downarrow$)}}\\ 
\cmidrule{2-5}
& Grid\_2$\times$2 & Grid\_3$\times$3 & Grid\_4$\times$4 & Grid\_5$\times$5 \\
\midrule
RGPE & 205.99 ± 1.41 & 564.22 ± 0.88 & 978.70 ± 4.20 & 1548.31 ± 3.18 \\
ABLR & 205.49 ± 0.64 & 561.79 ± 1.35 & 986.09 ± 1.53 & 1555.60 ± 4.49 \\
FSBO & 204.60 ± 2.27 & 563.54 ± 2.07 & 972.85 ± 7.68 & 1549.79 ± 0.40 \\
\bottomrule
Ours & \cellcolor{lightgray}\textcolor{black}{\textbf{197.38 ± 6.44}} & \cellcolor{lightgray}\textcolor{black}{\textbf{552.34 ± 2.55}} & \cellcolor{lightgray}\textcolor{black}{\textbf{923.45 ± 8.21}} & \cellcolor{lightgray}\textcolor{black}{\textbf{1519.29 ± 7.66}} \\
\bottomrule
\end{tabular}
}
\label{table:ablation2_2}
\end{table}
\begin{table}[t]
\centering
\caption{Ablation on different acquisition functions. Experiment is conducted on phase time allocation task.}
\vspace{-1pt}
\resizebox{1.0\linewidth}{!}{
\begin{tabular}{ccccc}
\toprule
\multirow{2}{*}{\textbf{Method}} & \multicolumn{4}{c}{\textbf{Average Number of Waiting Vehicles ($\downarrow$)}}\\ 
\cmidrule{2-5}
& Grid\_2$\times$2 & Grid\_3$\times$3 & Grid\_4$\times$4 & Grid\_5$\times$5 \\
\midrule
Ours (UCB) & 46.47 ± 0.55 & 146.94 ± 0.97 & 247.05 ± 2.48 & 421.74 ± 0.51\\
Ours (EI) & 46.99 ± 0.64 & 147.52 ± 1.04 & 247.89 ± 3.02 & 422.64 ± 0.51 \\
Ours (PI) & 46.33 ± 0.97 & 146.16 ± 0.83 & 246.47 ± 2.95 & 426.13 ± 0.53\\
\bottomrule
\end{tabular}
}
\label{table:ablation4}
\end{table}
\begin{table}[t]
\centering
\caption{Ablation on different NP architectures. Experiment is conducted on phase time allocation task.}
\vspace{-1pt}
\resizebox{1.0\linewidth}{!}{
\begin{tabular}{ccccc}
\toprule
\multirow{2}{*}{\textbf{Method}} & \multicolumn{4}{c}{\textbf{Average Number of Waiting Vehicles ($\downarrow$)}}\\ 
\cmidrule{2-5}
& Grid\_2$\times$2 & Grid\_3$\times$3 & Grid\_4$\times$4 & Grid\_5$\times$5 \\
\midrule
ANP & \cellcolor{lightgray}\textcolor{black}{\textbf{46.47 ± 0.55}} & \cellcolor{lightgray}\textcolor{black}{\textbf{146.94 ± 0.97}} & \cellcolor{lightgray}\textcolor{black}{\textbf{247.05 ± 2.48}} & \cellcolor{lightgray}\textcolor{black}{\textbf{421.74 ± 0.51}}\\
TNP-D & 47.25 ± 0.07 & 147.09 ± 0.96 & 255.68 ± 11.31 & 435.00 ± 1.99\\
\bottomrule
\end{tabular}
}
\vspace{-10pt}
\label{table:ablation5}
\end{table}

\subsubsection{\textbf{Evaluation on Different Metrics}}\label{app:add_exp2}
There are several different evaluation metrics to validate the efficiency of the proposed design on traffic system. We prepare two additional evaluation metrics: average traveling time and CO2 emissions. To compute both metrics, we use \texttt{getTravelingtime()} and \texttt{getCO2Emission()} method implemented by SUMO \citep{behrisch2011sumo}. 

We conduct experiments on synthetic grid environments with phase time allocation optimization task. As shown in the \Cref{table:ablation2_1,table:ablation2_2}, our method consistently outperforms other state-of-the-art baselines even when we change the evaluation metrics.


\subsubsection{\textbf{Acquisition Function Choices}}\label{app:add_exp4}
We investigate the dependency of our method in different acquisition functions during online adaption with Bayesian optimization. To validate this, we prepare two additional acquisition functions, expected improvement (EI) and probability of improvement (PI). 
As shown in the \Cref{table:ablation4}, our method demonstrates robust performance across various acquisition functions.

\subsubsection{\textbf{Model Architecture Choices}}\label{app:add_exp5}
We choose ANP as a meta-surrogate model for our method due to its simplicity and expressivity. We notice that Transformer Neural Processes (TNP, \cite{nguyen2022transformer}) have recently been proposed, which exhibit state-of-the-art performance in various benchmark problems. To this end, we compare our method with TNP-D as a meta-surrogate model. 

We perform our experiments on Grid 2$\times$2, 3$\times$3, 4$\times$4, and 5$\times$5 networks with phase time allocation task. We present the experiment results in the \Cref{table:ablation5}. Unlike experiment results in synthetic benchmark problems reported in recent papers, ANP shows better performance compared to TNP-D in our experiment setting. We speculate that this discrepancy occurs due to the size of the dataset. Unlike benchmark problems, we have a limited offline dataset that covers only a small manifold of the whole space. In these circumstances, expressive methods such as TNP-D might suffer from severe overfitting issues and lead to poor performance.

Note that our contribution is not developing a novel NP architecture but introducing a practical framework and method to solve real-world traffic problems. We hope that our research can raise the question of whether highly developing NP variants tested in specific benchmark problems are truly beneficial in real-world problems.

\begin{table}[t]
\centering
\caption{Comparison of the average number of passing vehicles on District A before and after applying our method.}
\vspace{-7pt}
\resizebox{\linewidth}{!}{
\begin{tabular}{ccccc}
\toprule
\multirow{2}{*}{\textbf{Method}} & \multicolumn{4}{c}{\textbf{Avg. Number of Passing Vehicles ($\uparrow$)}}\\
\cmidrule{2-5}
  & Tue & Wed & Thu & Overall\\
\midrule
Original Strategy & 16775 & 16072 & 19185 &\cellcolor{lightgray}\textcolor{black}{\textbf{17344}}\\
Proposed Method   & 17672 & 17735 & 19120 &\cellcolor{lightgray}\textcolor{black}{\textbf{18176}}\\
\midrule
Improvement (\%)	&	5.35	& 10.35 & -0.34 &\cellcolor{lightgray}\textcolor{black}{\textbf{4.80}}\\
\bottomrule
\end{tabular}}
\label{table:deployment}
\vspace{-10pt}
\end{table}
\begin{figure*}[t]
    \centering
    \vspace{-7pt}
    \includegraphics[width=\linewidth]{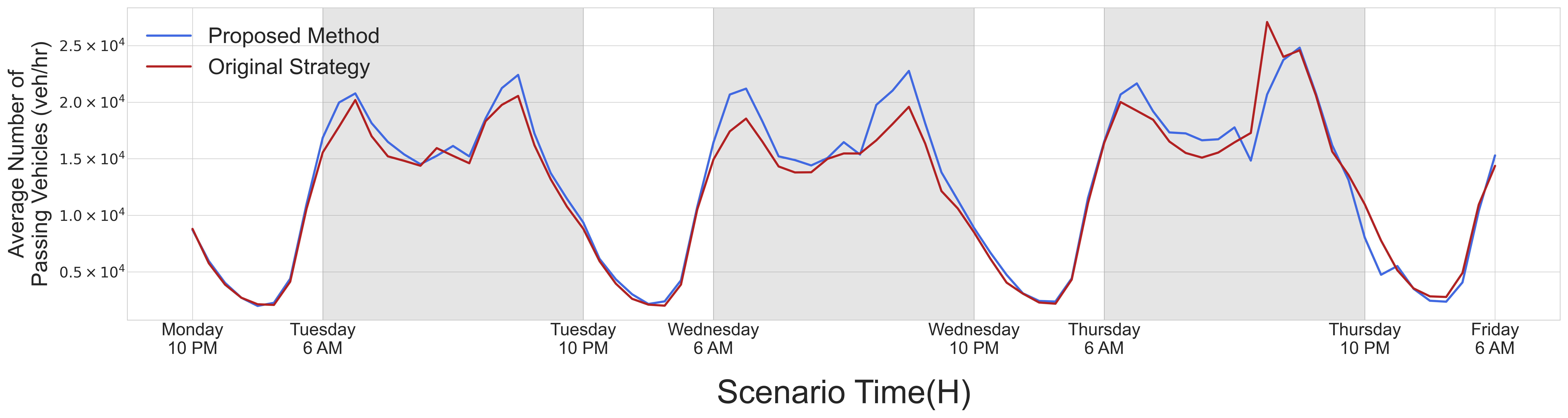}
    \vspace{-20pt}
    \caption{Performance of proposed method in real-world deployment.}
    \label{fig:deployment}
\end{figure*}

\section{Real-World Deployment}
We implement the optimized phase time allocation design for the 26 actual traffic lights. Due to privacy issues, we anonymize the deployment site as District A. Please refer to \Cref{app:deployment} for more information about the deployment.

For deployment, we initiate by collecting real traffic data using surveillance cameras, spanning from 6 AM to 10 PM for five weeks. We assume that the traffic pattern differs per hour. Subsequently, we train the ANP model using data from the first four weeks and find an optimal phase time allocation design for each hour of the last week. Finally, we apply the derived phase time allocation for the last week of the offline data right into the real-world traffic system. Note that because we cannot compare different operating systems in the completely same environment, we choose the week from the same month that has the most similar traffic pattern.

\Cref{table:deployment} compares the performance of the proposed method with the original fixed-time strategy used in practice. As shown in the table, our method improves the overall throughput by an average of 4.80\% compared with the original fixed-time strategy. \Cref{fig:deployment} shows the overall performance of our method compared to the original fixed-time strategy across three days. Our method improves the traffic flow most of the time except few hours. 
It should be noted that our method has manifested substantial improvement in the real traffic system despite its low implementation cost.

\section{Related Works}

\subsection{Traffic Light Optimization}
Optimizing multiple traffic lights in urban road networks to alleviate traffic congestion has been widely studied. \citet{zhang2010traffic, li2018signal} apply genetic algorithm to determine the optimal phase time allocation while \citet{dong2010urban, dabiri2016arterial} introduce particle swarm optimization to optimize the phase time allocation for the overall traffic network. Recently, \citet{tay2022bayesian} propose a Bayesian optimization-based approach tailored for large-scale traffic networks. 
These methods effectively find appropriate traffic light designs but possess limitations. They require extensive simulation-based evaluations, initiating the optimization process from scratch for distinct traffic patterns. While our method aligns with the literature by formulating the design optimization as a black-box optimization, it distinguishes itself by leveraging prior knowledge from previously collected offline meta dataset for fast adaptation to unseen traffic patterns.

\subsection{Meta Black-box Optimization}
To overcome the aforementioned limitations, meta-learning based black-box optimization has been proposed. \citet{tjarko2022discovering} meta-learn to recombine solution candidates in CMA-ES via dot-product self-attention. Similarly, \citet{lange2023discovering} parameterize selection and mutation rate in genetic algorithms as cross- and self-attention modules and train them on various tasks. 

There is also a line of research that tries to make extensions of GP to facilitate meta black-box optimization. \citet{feurer2018scalable} utilize a weighted combination of independent GP models and adjust weight during online adaptation. \citet{perrone2018scalable} utilize a shared feature map coupled with an adaptive Bayesian linear regression layer for each task. \citet{wistuba2021few} cast meta Bayesian optimization as a few-shot learning problem and train initial deep kernel parameters via model-agnostic meta-learning \citep{finn2017model}. However, a majority of the meta-learning based black-box optimization methods actively utilize online evaluations for various tasks, which is not practical for real-world systems. Contrasting with prior methods, our approach harnesses an offline meta dataset for training a meta prediction model with well-calibrated uncertainty to prevent distributional shift issue during online adaptation process \cite{fujimoto2019off, kumar2019stabilizing}.
\section{Conclusion}
In this study, we aim to derive an efficient urban traffic light management system by solving offline meta black-box optimization problems on two pivotal design components: phase combination and phase time allocation. We propose a novel combination of ANP and BO which can identify underlying similarities between traffic patterns and transfer the knowledge effectively from the offline meta dataset to the online adaptation process.
We demonstrate the superiority of our novel combination not only on simulation-based various urban road networks with diverse traffic patterns but also on real-world deployment.

\begin{acks}
This work was supported by Institute of Information \& communications Technology Planning \& Evaluation (IITP) grant funded by the Korea government(MSIT) (2022-0-01032, Development of Collective Collaboration Intelligence Framework for Internet of Autonomous Things)
\end{acks}
\clearpage

\bibliographystyle{ACM-Reference-Format}
\bibliography{references}

\appendix
\begin{table*}[t]
\centering
\caption{Model hyperparameters}
\vspace{-7pt}
\resizebox{0.85\linewidth}{!}{
\begin{tabular}{cccccccc}
\toprule
  & Grid\_2$\times$2 & Grid\_3$\times$3 & Grid\_4$\times$4 & Grid\_5$\times$5 & Hangzhou\_4$\times$4 & Manhattan\_16$\times$3 & Manhattan\_28$\times$7\\
\midrule
Number of Encoder Layers & 2 & 2 & 2 & 2 & 2 & 3 & 3\\
Number of Decoder Layers & 2 & 2 & 2 & 2 & 2 & 3 & 3\\
Hidden Units & 32 & 32 & 32 & 32 & 32 & 64 & 64\\
Training Steps & 10K & 10K & 10K & 10K & 10K & 25K & 25K\\
\bottomrule
\end{tabular}}
\label{table:hyperparam}
\end{table*}
\section{Experiment Details}
\subsection{Environment Setup}\label{app:exp_setup}
In this subsection, we provide more details of our experiment setup. Note that we provide our code for experiments in \url{https://anonymous.4open.science/r/offline_meta_bbo-4DE4/}.

\subsubsection{Traffic Networks}
For all networks, each road consists of three lanes. Each lane is designated for a specific direction of vehicle movement: vehicles can choose lane among turning left, going straight and turning right. For synthetic grid networks, each intersection is separated by a distance of 300m. Real-world road networks are imported from OpenStreetMap\footnote{https://www.openstreetmap.org/}.

\subsubsection{Traffic Flow}
For synthetic road networks, we pre-define origin-destination (OD) pairs which are mainly composed of four directions (North $\leftrightarrow$ South, East $\leftrightarrow$ West). To create diverse patterns, we randomly assign the arrival rate for each OD pair. Concretely, we use the Poisson Process with the parameter $\lambda$, where $\lambda$ represents the expected value of generated vehicles per second (arrival rate). Then the time interval between the generation of vehicles, denoted by $\Delta t$, follows the exponential distribution with $\lambda$. For each task, we randomly generate parameters to satisfy $\lambda_{NS}+\lambda_{SN}+\lambda_{EW}+\lambda_{WE}=0.1$ and $0.01 
\leq \lambda_{NS}, \lambda_{SN}, \lambda_{EW}, \lambda_{WE}\leq0.1$, where $\lambda_{NS}$ represents the Poisson parameters from North to South, and $\lambda_{SN}$, $\lambda_{EW}$, and $\lambda_{WE}$ are defined in the same manner. By sampling parameters in this way, we make the number of total vehicles appeared in the traffic network similar across different traffic patterns.

For real-world networks, we already have data collected from the real-world traffic. To create diverse traffic patterns with this data, we do the following procedure. First, we sort vehicle movements by departure time. Then, we randomly shuffle the departure time of vehicle movements so that the total number of vehicles that appeared during the simulation is preserved as follows:
\begin{align}
    &\text{Original Traffic Pattern:} \{(o_1, d_1, t_1), \cdots, (o_V, d_V, t_V)\}\\
    &\text{New Traffic Pattern:} \{(o_1, d_1, t_{\sigma(1)}), \cdots, (o_V, d_V, t_{\sigma(V)})\}
\end{align}
where $\sigma(\cdot)$ is a permutation operator.

\subsubsection{Congestion Measure}
For all experiments, we simulate the proposed design of traffic lights under a specific traffic pattern and network for 30 minutes in simulation time. To compute the average number of waiting vehicles, we monitor the overall traffic network every 60 seconds and compute the sum of the number of waiting vehicles for all intersections at that time. Then, we report the mean of the sum collected from every interval.

\subsection{Model Hyperparameters}\label{app:hyperparameters}
For all experiments, we fix the learning rate as 0.005 and use Adam \citep{kingma2014adam} optimizer for updating parameters. We checkpoint the model every 200 training epochs and save the model with the highest target log-likelihood on the validation dataset. \Cref{table:hyperparam} shows the hyperparameters we used for various tasks.

\subsection{Task Details}\label{app:task_details}
\subsubsection{Phase Combination}
For the phase combination task, our decision variable $\mathbf{x}=[x_1, x_2, \cdots, x_I]$ depicts the phase combination of all intersections in a traffic network, where $x_i$ indicates the phase combination of $i$th intersection. Specifically, $x_i$ can be defined as a vector, $[x_{i1},\cdots,x_{iJ}]$, where $J$ is the number of possible phase combinations and $x_{ij}\in[0, 1]$ is the unnormalized probability to choose $j$th phase combination for $i$th intersection. We select the phase combination for $i$th intersection among $J$ combinations by choosing the index with maximum value from the decision variable as follows:
\begin{align}
    \arg\max_{j}\frac{\exp(x_{ij})}{\sum_{k=1}^{J}\exp(x_{ik})}\quad\forall i=1,\cdots,I
\end{align}

\subsubsection{Phase Time Allocation}
For the phase time allocation task, our decision variable $\mathbf{x}=[x_1, x_2, \cdots, x_I]$ depicts the phase allocation of all intersections in a traffic network, where $x_i$ indicates the phase allocation plan of $i$th intersection. Specifically, $x_i$ can be defined as a vector, $[x_{i1}, \cdots, x_{iJ}]$, where $J$ is the number of phases in phase combination selected by $i$th intersection and $x_{ij}\in[0, 1]$ is the unnormalized ratio assigned to $j$th phase of $i$th intersection for all $j=1,\cdots,J$. Given cycle time $c$ and minimum green time for each phase $g_{\text{min}}$, we compute the real allocation time for $j$th phase of $i$th intersection from the decision variable as follows:
\begin{align}
    g_{\text{min}} + (c-g_{\text{min}}\cdot J)\cdot\frac{\exp(x_{ij})}{\sum_{k=1}^{J}\exp(x_{ik})}\quad\forall i=1,\cdots,I\:j=1,\cdots,J
\end{align}
\section{Bayesian Optimization Details}\label{app:bo_details}

In this section, we describe details on Bayesian optimization, especially on acquisition functions that we use in our experiments. Acquisition functions are used to balance predicted mean $\mu(\mathbf{x})$ and variance $\sigma^2(\mathbf{x})$ from the surrogate model of BO to find an optimal-looking candidate.

\textit{Upper Confidence Bound}. The upper confidence bound (UCB) acquisition function is defined as:
\begin{align}
    \text{UCB}(\mathbf{x})=\mu(\mathbf{x}) + \beta\cdot\sigma(\mathbf{x})
\end{align}
where $\beta > 0$ is a tunable parameter that controls the trade-off between exploration and exploitation. For all experiments, we set $\beta$ as 2.0.

\textit{Expected Improvement}.  The expected improvement is defined over the current best-observed value $f(\mathbf{x}^{+})$ as follows:
\begin{align}
    \text{EI}(\mathbf{x})=(\mu(\mathbf{x})-f(\mathbf{x}^{+}))\cdot\Phi(\mathbf{Z}) + \sigma(\mathbf{x})\cdot\phi(Z)
\end{align}
where $\Phi(Z)$ and $\phi(\mathbf{Z})$ are cumulative and probability density function of standard normal distribution, respectively.

\textit{Probability of Improvement}. The probability of improvement (PI) is defined over the current best-observed value $f(\mathbf{x}^{+})$ as follows:
\begin{align}
    \text{PI}(\mathbf{x})=\Phi\left(\frac{\mu(\mathbf{x})-f(\mathbf{x}^{+})}{\sigma(\mathbf{x})}\right)
\end{align}
where $\Phi(\cdot)$ is a cumulative density function of standard normal distribution.

While there are other acquisition functions such as Thompson Sampling, Entropy Search \citep{hennig2012entropy} and Predictive Entropy Search \citep{hernandez2014predictive}, we experiment with the above three most widely used acquisition functions for our experiments. For more details on acquisition functions, please refer to this paper \citep{frazier2018tutorial}.

\section{Pseudocode for our proposed method}
The pseudocode of our algorithm is described in \Cref{alg:pseudocode}.

\begin{algorithm}
    \caption{Proposed Method}
    \begin{algorithmic}
    \Require{\\$\mathcal{M}_{\theta}$ - Neural Process model
             \\$\mathcal{A}$ - Acquisition function
             \\$f^{1}, f^{2}, \dots, f^{N}$ - Similar traffic patterns
             \\$f^{\text{unseen}}$ - Unseen traffic pattern
             \\$K$ - Maximum number of trials for Bayesian Optimization}
    \\
    \State{// Phase 1: Data Collection}
    \State{Initialize the meta dataset $\mathcal{D}=\{\}$}
    \For {$n=1,\cdots,N$}
        \State{Initialize the dataset $\mathcal{D}^{n}=\{\}$}
        \For {$m=1,\cdots,M$}
            \State{Random sample the candidate $\mathbf{x}_m^n \in \mathcal{X}$}
            \State{Evaluate target function $y_m^n=f^n(\mathbf{x}_m^n$)}
            \State{Update $\mathcal{D}^n\leftarrow\mathcal{D}^{n}\cup\{(\mathbf{x}_{m}^n, y_m^n)\}$}
        \EndFor
        \State{Update $\mathcal{D}\leftarrow\mathcal{D}\cup\{\mathcal{D}^n\}$} 
    \EndFor
    
    \\
    \State{// Phase 2: Meta-train $\mathcal{M}_{\theta}$}
    \While {$\theta$ converges}
        \State{Sample a task: $n\in\{1,\cdots,N\}$}
        \State{Sample context and target set: $\mathcal{D}^{n}_{\text{ctx}}, \mathcal{D}^{n}_{\text{tar}}\sim\mathcal{D}^{n}$}
        \State{Update parameters $\theta$ to maximize Equation \ref{eq:elbo}}
    \EndWhile
    \\
    \State{// Phase 3: Perform Bayesian Optimization}
    \State{Initialize the online dataset $\mathcal{D}^{\text{on}}=\{\}$}
    \For {t=$1,\cdots,T$}
        \State {Use $\mathcal{D}^{\text{on}}$ as a context set for $\mathcal{M}_{\theta}$}
        \State {Find $\mathbf{x}_{k}$ that maximizes $\mathcal{A}(\mathbf{x};\mathcal{M}_{\theta})$}
        \State {Evaluate $y_{k}=f^{\text{unseen}}(\mathbf{x}_{k})$}
        \State {Update $\mathcal{D}^{\text{on}}\leftarrow\mathcal{D}^{\text{on}}\cup\{(\mathbf{x}_{k}, y_k)\}$}
    \EndFor
    \\
    \Return {$\arg\max_{\mathbf{x}\in\mathcal{D}^{\text{on}}}f^{\text{unseen}}(\mathbf{x})$}
    \end{algorithmic}
    \label{alg:pseudocode}
\end{algorithm}
\section{Real-world Deployment}\label{app:deployment}

We control 26 heterogeneous intersections in district A to evaluate our method. Each intersection has a different number of phases, cycle time, and minimum green time for each phase.


We collect real-world traffic data from smart surveillance cameras, similar to the Hangzhou dataset. The main difference from the public dataset is that we have traffic data with various traffic patterns. We create different sets of vehicle movements using traffic data gathered from different timezones. Due to confidentiality agreements with the company that provides the data, we are afraid that the dataset and simulator used in this study cannot be made publicly available.

\section{Discussion and Future Work}
We are interested in extending our method. 
One can consider multi-objective optimization methods \citep{wang2023high} to achieve improvements on various evaluation metrics. Furthermore, there is still room for additional variables affecting the efficiency of traffic lights, such as offset and cycle time length. While our method can be directly applied to those variables, methods that carefully consider the properties of different variables seem more promising. 

\clearpage

\end{document}